\documentclass{article}

\usepackage{arxiv}

\usepackage[utf8]{inputenc} % allow utf-8 input
\usepackage[T1]{fontenc}    % use 8-bit T1 fonts
\usepackage{hyperref}       % hyperlinks
\usepackage{url}            % simple URL typesetting
\usepackage{booktabs}       % professional-quality tables
\usepackage{amsfonts}       % blackboard math symbols
\usepackage{nicefrac}       % compact symbols for 1/2, etc.
\usepackage{microtype}      % microtypography
\usepackage[pdftex]{graphicx}
\usepackage{amssymb,amsfonts,amsmath,amscd}
\usepackage[printonlyused,withpage]{acronym}
\usepackage{natbib}
\usepackage{mathrsfs}
\usepackage{breqn}

\graphicspath{{figs/}}
\newcommand{\bbm}{\begin{bmatrix}}
\newcommand{\ebm}{\end{bmatrix}}
\newcommand{\mbf}{\mathbf}
\newcommand{\mbs}[1]{{\boldsymbol{#1}}}

\newcommand{\beq}{\begin{equation}}
\newcommand{\eeq}{\end{equation}}
\newcommand{\bdis}{\begin{displaymath}}
\newcommand{\edis}{\end{displaymath}}
\newcommand{\beqn}[1]{\begin{subequations}\label{eq:#1}\begin{eqnarray}}
\newcommand{\eeqn}{\end{eqnarray}\end{subequations}}
\newcommand{\wdg}{\wedge}

\acrodef{BA}{Bundle Adjustment}
\acrodef{DNN}{Deep Neural Network}
\acrodef{EM}{Expectation Maximization}
\acrodef{GEM}{Generalized Expectation Maximization}
\acrodef{HMM}{Hidden Markov Model}
\acrodef{LDS}{Linear Dynamical System}
\acrodef{LQG}{Linear Quadratic Gaussian}
\acrodef{LQR}{Linear Quadratic Regulator}
\acrodef{LTI}{Linear Time-Invariant}
\acrodef{RTS}{Rauch-Tung-Striebel}
\acrodef{SGD}{Stochastic Gradient Descent}
\acrodef{SLAM}{Simultaneous Localization and Mapping}
\acrodef{RKHS}{Reproducing Kernel Hilbert Space}
\acrodef{SMW}{Sherman-Morrison-Woodbury}

\acrodef{GVI}{Gaussian Variational Inference}
\acrodef{ESGVI}{Exactly Sparse Gaussian Variational Inference}
\acrodef{MAP}{Maximum A Posteriori}
\acrodef{ML}{Maximum Likelihood}
\acrodef{KL}{Kullback-Leibler}
\acrodef{PDF}{Probability Density Function}
\acrodef{NEES}{Normalized Estimation Squared Error}
\acrodef{KF}{Kalman Filter}
\acrodef{VKF}{Variational Kalman Filter}
\acrodef{ISPKF}{Iterated Sigmapoint Kalman Filter}
\acrodef{ESGVI-GN}{ESGVI Gauss-Newton}
\acrodef{ELBO}{Evidence Lower Bound}
\acrodef{NGD}{Natural Gradient Descent}
\acrodef{FIM}{Fisher Information Matrix}
\acrodef{RANSAC}{Random Sample And Consensus}
\acrodef{IRLS}{Iteratively Reweighted Least-Squares}
\acrodef{BRD}{Black-Rangarajan Duality}
\acrodef{GNC}{Graduated Non-Convexity}
\acrodef{GA}{Geometric Algebra}

\hypersetup{%
    pdftitle={},
    pdfauthor={},
    pdfkeywords={},
    pdfsubject={},
    pdfstartview=FitH,%
    bookmarks=true,%
    bookmarksopen=true,%
    breaklinks=true,%
    colorlinks=true,%
    linkcolor=black,
    anchorcolor=black,%
    citecolor=black,
    filecolor=black,%
    menucolor=black,
    pagecolor=black,%
    urlcolor=black
}%

\title{A Geometric Algebra Solution to Wahba's Problem}

\author{
 \normalfont Timothy D. Barfoot \\
 Institute for Aerospace Studies \\
 University of Toronto \\
 \texttt{tim.barfoot@utoronto.ca} 
}

\begin{document}

\maketitle

\begin{abstract}
We retrace Davenport's solution to Wahba's classic problem of aligning two pointclouds using the formalism of \ac{GA}.  \ac{GA} proves to be a natural backdrop for this problem involving three-dimensional rotations due to the isomorphism between unit-length quaternions and {\em rotors}.  While the solution to this problem is not a new result, it is hoped that its treatment in \ac{GA} will have tutorial value as well as open the door to addressing more complex problems in a similar way.
\end{abstract}


\section{Geometric Algebra Background}

\acf{GA} has been touted as a `universal mathematical language' that has been used with great success to synthesize well-known results in mathematics, physics, and several other applications.  \ac{GA} was first concieved by \citet{clifford78} and has often gone by the name {\em Clifford algebra}; notably, it unifies quaternions \citep{hamilton66} and the exterior (related to the cross) product \citep{grassmann44}, both key tools in describing three-dimensional space.  \ac{GA} lay dormant until rediscovered by Hestenes in the 1960s \citep{hestenes66, hestenes17} who showed its power to describe classical, electromagnetic, and even quantum physics.  \citet{bayro18,bayro20} provides an extensive overview of \ac{GA} applications outside of physics, notably in computer vision and robotics.

\subsection{Geometric Product and Multivectors}

As part of our goal is to be tutorial, we introduce just enough of \ac{GA} to get a taste of its ingenuity, while not overwhelming the reader with mathematical formalism.  We also refer to \citet{miller13}, who provides a gentle introduction to \ac{GA} and whose notation we mimic.  We will restrict ourselves to one particular geometric algebra, $\mathbb{G}^3$, which is actually the simplest but also least powerful way of describing three-dimensional space.  There are many different geometric algebras with different properties that can be used to describe more complex geometry than we need here \citep{bayro18,bayro20}.

The novel aspect of \ac{GA} is that it allows us to multiply two vectors, $\mbf{u}$ and $\mbf{v}$, by defining the {\em geometric product}:
\begin{equation}
\mbf{u} \mbf{v} = \underbrace{\mbf{u}\cdot \mbf{v}}_{\rm scalar} + \underbrace{\mbf{u} \wdg \mbf{v}}_{\rm bivector},
\end{equation}
where $\cdot$ indicates the usual inner (or dot) product and $\wdg$ indicates the outer (or exterior) product.  This is quite interesting because we see already that we are adding together two things that would normally not be allowed in matrix algebra, a scalar and a new quantity called a {\em bivector}.  

A bivector is formed through the outer product of two vectors; in three dimensions, it is very similar to the cross product, but actually works in any number of dimensions.  The cross product produces a vector that is normal to the plane in which the two vectors lie;  the outer product, $\mbf{u} \wdg \mbf{v}$, creates an {\em oriented area}, which is a parallelogram in the plane formed by the two vectors along with a binary variable indicating which side is `up'.  
The outer product enjoys some other properties similar to the cross product:
\begin{equation}
\mbf{u} \wdg \mbf{v} = - \mbf{v} \wdg \mbf{u}, \quad \mbf{u} \wdg \mbf{u} = \mbf{0}.
\end{equation}
So far, we have the usual scalars, vectors, and our new bivectors.  For our description of three-dimensional space, we can also have a {\em trivector}, which is simply the (associative) geometric product of three vectors.  We refer generically to the addition of these different types of objects as a {\em multivector}:
\begin{equation}
\mbs{M} = \underbrace{a}_{\rm scalar} + \underbrace{\mbf{b}}_{\rm vector} + \underbrace{\mbs{C}}_{\rm bivector} + \underbrace{\mbs{D}}_{\rm trivector}.
\end{equation}
The addition of different types of objects may seem unusual at first, but this is a familiar concept when using complex numbers, $a + i b$, for example; in fact, \ac{GA} is a more geometrically understandable way of thinking about complex numbers.

In the vectorspace $\mathbb{R}^3$, we can pick an orthonormal basis, $\{ \mbf{e}_1, \mbf{e}_2, \mbf{e}_3 \}$.  Owing to orthonormality, we have
\begin{equation}
\mbf{e}_i\mbf{e}_j = \left\{ \begin{array}{cl}  1 & i = j \\ \mbf{e}_i \wdg \mbf{e}_j & i \neq j \end{array}  \right. .
\end{equation}
Since $\mbf{e}_i \mbf{e}_j = - \mbf{e}_j \mbf{e}_i$ for $i \neq j$, we have three linearly independent {\em unit bivectors}, $\{ \mbf{e}_2 \mbf{e}_3, \mbf{e}_3 \mbf{e}_1, \mbf{e}_1 \mbf{e}_2 \}$.  There is only one linearly independent {\em unit trivector}, $\mbf{e}_1 \mbf{e}_2 \mbf{e}_3$; this is sometimes referred to as a {\em pseudoscalar} and is given a unique symbol,
\begin{equation}
\mbs{I} = \mbf{e}_1 \mbf{e}_2 \mbf{e}_3.
\end{equation}
Notably, we have
\begin{equation}
\mbs{I}^2 = -1,
\end{equation}
which is related to the usual $i^2 = -1$ from complex numbers.  Another useful property is that
\begin{equation}
\mbs{I} \mbf{v} = \mbf{v} \mbs{I},
\end{equation}
for vector $\mbf{v}$.
Finally, in three dimensions we can relate the outer and cross products using $\mbs{I}$ according to
\begin{equation}
\underbrace{\mbf{u} \wdg \mbf{v}}_{\rm bivector} = \mbs{I} \underbrace{\mbf{u}^\wdg \mbf{v}}_{\rm vector}
\end{equation}
where the skew-symmetric matrix operator, $\wdg$, that implements the cross product is defined as
\begin{equation}
\mbf{u}^\wdg = \bbm u_1 \\ u_2 \\ u_3 \ebm^\wdg = \bbm 0 & -u_3 & u_2 \\ u_3 & 0 & -u_1 \\ -u_2 & u_1 & 0 \ebm.
\end{equation}
Comfortingly, the $\wdg$ symbol that is used frequently to express both the geometric and cross products finds common meaning here.

This means any multivector can be written as a linear combination of all the unit multivectors as follows:
\begin{equation}
\mbs{M} = a + b_1 \mbf{e}_2 + b_2 \mbf{e}_2 + b_3 \mbf{e}_3 + c_1 \mbf{e}_2\mbf{e}_3 + c_2 \mbf{e}_3\mbf{e}_1 + c_3 \mbf{e}_1\mbf{e}_2 + d \mbf{e}_1 \mbf{e}_2 \mbf{e}_3,
\end{equation}
for the eight scalar coefficients, $\{a, b_1, b_2, b_3, c_1, c_2, c_3, d\}$, or more compactly as
\begin{equation}\label{eq:multivector}
\mbs{M} = a + \mbf{b} + \mbs{I} \mbf{c} + \mbs{I} d,
\end{equation}
where $\mbf{b} = (b_1, b_2, b_3)$ and $\mbf{c} = (c_1, c_2, c_3)$.  We can therefore define a new eight-dimensional vectorspace, $\mathbb{G}^3$, as
\begin{equation}
\mathbb{G}^3 = \mbox{span}\{ 1, \mbf{e}_2, \mbf{e}_2, \mbf{e}_3, \mbf{e}_2\mbf{e}_3, \mbf{e}_3\mbf{e}_1, \mbf{e}_2 \mbf{e}_3, \mbf{e}_1 \mbf{e}_2\mbf{e}_3 \},
\end{equation}
which consists of all linear combinations of these eight basis multivectors over the field, $\mathbb{R}$.  The usual properties of a vectorspace can be easily verified (i.e., closure under addition, commutativity, zero element, and so on).  

Although we discussed applying the geometric product to two vectors, we can easily apply it to any two multivectors in $\mathbb{G}^3$ and the result will be a multivector also in $\mathbb{G}^3$:
\begin{equation}
\mbs{M}_1 \mbs{M}_2 \in \mathbb{G}^3,
\end{equation}
with $\mbs{M}_1, \mbs{M}_2 \in \mathbb{G}^3$. It thus turns out that we can extend $\mathbb{G}^3$ to an {\em algebra} under the geometric product since the usual properties can also be verified (i.e., associativity, identity element, inverse, closure under the product, and so on); therefore, $\mathbb{G}^3$ is referred to as a {\em geometric algebra}.  

We will need a few more operations when working with geometric algebra.  The {\em reverse} of a multivector, $\widetilde{\mbs{M}}$, is computed by reversing the order of any vectors in geometric products, which for $\mathbb{G}^3$ results in
\begin{equation}
\widetilde{\mbs{M}} = a + \mbf{b} - \mbs{I} \mbf{c} - \mbs{I} d,
\end{equation}
where $\mbs{M}$ is defined as in~\eqref{eq:multivector}.
The {\em scalar projection} of a multivector (onto the basis $\{1\}$) is defined using the $\left< \cdot \right>$ operation:
\begin{equation}
\left< \mbs{M} \right> = a,
\end{equation}
with $\mbs{M}$ again defined as in~\eqref{eq:multivector}.  This operation obeys a cyclic property:
\begin{equation}
\left< \mbs{M}_1 \mbs{M}_2 \cdots \mbs{M}_{K-1} \mbs{M}_K \right> = \left< \mbs{M}_2 \mbs{M}_3 \cdots \mbs{M}_K \mbs{M}_1 \right>,
\end{equation}
with $(\forall k) \, \mbs{M}_k \in \mathbb{G}^3$.  

\subsection{Even Subalgebra and Rotors}

A {\em subalgebra} is simply a subset of an algebra that enjoys all the same properties as its parent.  For $\mathbb{G}^3$, the following {\em even subalgebra} exists: 
\begin{equation}
\mathbb{G}^{3^+} = \mbox{span}\{ 1, \mbf{e}_2\mbf{e}_3, \mbf{e}_3\mbf{e}_1, \mbf{e}_2 \mbf{e}_3 \},
\end{equation} 
which consists of all linear combinations of the unit scalar and unit bivector basis multivectors; it is refered to as `even' since it keeps only the basis multivectors that can be written as the geometric product of an even (not odd) number of vectors (i.e., $0$ and $2$, not $1$ and $3$).  Clearly $\mathbb{G}^{3^+}$ is a (four-dimensional) subspace of $\mathbb{G}^{3}$ and it can be readily verified that it is a subalgebra as well (e.g., geometric product of two elements remains in the subalgebra).  Every multivector in $\mathbb{G}^{3^+}$ can be written as
\begin{equation}\label{eq:evenmultivector}
\mbs{M}^+ = a + \mbs{I} \mbf{c},
\end{equation}
for coefficients $a$ and $\mbf{c}$.  The geometric product of two such even multivectors is
\begin{equation}\label{eq:evencompound}
\mbs{M}_1^+ \mbs{M}_2^+ = \left( a_1 + \mbs{I} \mbf{c}_1 \right) \left( a_2 + \mbs{I} \mbf{c}_2 \right) = \underbrace{\left( a_1 a_2 - \mbf{c}_1 \cdot \mbf{c}_2 \right)}_{\rm scalar} + \mbs{I} \underbrace{\left( a_1 \mbf{c}_2 + a_2 \mbf{c}_1 - \mbf{c}_1^\wdg \mbf{c}_2 \right)}_{\rm vector},
\end{equation}
which we see is still in $\mathbb{G}^{3^+}$.  The reverse of an even multivector is also an even multivector:
\begin{equation}
\widetilde{\mbs{M}}^+ = a - \mbs{I} \mbf{c}, 
\end{equation}
where $\mbs{M}^+$ defined as in~\eqref{eq:evenmultivector}.  The following identity is also quite useful for even multivectors:
\begin{equation}\label{eq:evenflip}
\mbs{M}^+ \mbf{v} = \mbf{v} \widetilde{\mbs{M}}^+ + 2 (\mbf{c} \cdot \mbf{v}) \mbs{I},
\end{equation}
where $\mbf{v}$ is a vector and $\mbs{M}^+$ defined as in~\eqref{eq:evenmultivector}.

Elements of the even subalgebra, $\mbs{M}^+ = a + \mbs{I} \mbf{c} \in \mathbb{G}^{3^+}$, that have the constraint $a^2 + \mbf{c} \cdot \mbf{c} = 1$ are called {\em rotors} and can be used to represent rotations.   Every rotor, $\mbs{R}$, can be written in the form
\begin{equation}\label{eq:rotor}
\mbs{R} = \cos \frac{\phi}{2} +  \sin \frac{\phi}{2} \mbs{I} \mbf{a} = \exp\left( \phi \mbs{I} \mbf{a} \right),
\end{equation}
where $\phi$ is an angle of rotation and $\mbf{a}$ is the axis of rotation; this can be viewed as the \ac{GA} version of Euler's formula.  The compounding of two rotors can be achieved using~\eqref{eq:evencompound} and the result can be shown to also be a rotor.  It turns out that the set of all rotors is not a subspace, but it is a {\em Lie group}, which has implications for optimization problems involving such quantities.  In fact, the set of rotors is isomorphic to the set of {\em unit-length quaternions}, a common representation of rotation.

The reverse of a rotor, $\mbs{R} = a + \mbs{I} \mbf{c}$, is 
\begin{equation}
\widetilde{\mbs{R}} = a - \mbs{I} \mbf{c},
\end{equation}
which is also the rotor's {\em geometric inverse}:  
\begin{equation}
\mbs{R} \widetilde{\mbs{R}} = \widetilde{\mbs{R}}  \mbs{R} = a^2 + \mbf{c} \cdot \mbf{c} = 1.
\end{equation}
The mechanics of using a rotor are as follows.  If we have a vector, $\mbf{v}$, we can rotate it through an angle $\phi$ about an axis $\mbf{a}$ according to
\begin{equation}
\mbf{v}^\prime = \mbs{R} \mbf{v} \widetilde{\mbs{R}},
\end{equation}
where $\mbf{v}^\prime$ is the rotated vector and $\mbs{R}$ is defined in~\eqref{eq:rotor}; this is similar to how unit-length quaternions are used to rotate vectors.  

\subsection{Geometric Calculus}

A calculus can also be defined for \ac{GA}, which is referred to as {\em geometric calculus}.  We will have need to take the derivative of scalar expressions with respect to elements of $\mathbb{G}^{3^+}$, the even subalgebra discussed in the previous section.  The {\em even multivector derivative} is defined \citep{lasenby98,doran01} to be
\begin{equation}
\frac{\partial}{\partial\mbs{M}^+} = \frac{\partial}{\partial a} - \sum_{i=1}^3 \mbs{I} \mbf{e}_i \frac{\partial}{\partial c_i},
\end{equation}
where $\mbs{M}^+ = a + \mbs{I} \mbf{c} \in \mathbb{G}^{3^+}$.

Under this definition, we have the useful results
\beqn{multideriv}
\frac{\partial}{\partial\mbs{M}_1^+} \left< \mbs{M}_1^+ \mbs{M}_2^+ \right> & = & \mbs{M}_2^+, \\
\frac{\partial}{\partial\mbs{M}_1^+} \left< \widetilde{\mbs{M}}_1^+ \mbs{M}_2^+ \right> & = & \widetilde{\mbs{M}}_2^+,
\eeqn
where $\mbs{M}_1^+, \mbs{M}_2^+ \in \mathbb{G}^{3^+}$.

\section{Wahba's Problem}

Wahba's problem \citep{wahba65} is a classic estimation problem involving three-dimensional rotations.  It has been extensively treated in the literature using both quaternions \citep{davenport65,shuster81,horn87,mortari97,barfoot_aa11,yang13,barfoot_ser17} and rotation matrices \citep{green52,horn87b,markley88,deruiter14}.  There have been geometric algebra solutions previously published \citep{lasenby98,doran01,bayro20} that culminate in solving a singular value decomposition problem.  Here we retrace the quaternion approach of \citet{davenport65} instead using \ac{GA} rotors, where the rotor constraint is enforced using a Lagrange multiplier term resulting in an eigenproblem.  Given the isomorphism between rotors and quaternions, this is a natural way to approach the problem in geometric algebra.  We also treat the common extension of the classic problem of Wahba by including unknown translation as well as rotation in our setup \citep{umeyama91}.

\subsection{Setup}

A {\em pointcloud}, is a collection of three-dimensional points expressed in a common reference frame.   One of the most fundamental problems in computer vision and state estimation is to find the three-dimensional relative translation and rotation between two such noisy pointclouds, $\mbf{u}_i$ and $\mbf{v}_i$.  We assume the points have already been associated with one another so that we know which point in one frame corresponds to which point in the other; we assume no incorrect matches.  

We would like to solve the following problem:
\begin{equation}\label{eq:prob1}
\min_{\mbs{R},\mbf{t}}\sum_i w_i \left( \mbf{u}_i - \mbs{R} (\mbf{v}_i - \mbf{t} )\widetilde{\mbs{R}} \right)^2
\end{equation}
with $\mbs{R}$ a rotor representing rotation, $\mbf{t} \in \mathbb{R}^3$ a vector representing translation, and $w_i$ some positive weights.  The first thing we will do is define a change of variables for the translation variable.  Let
\begin{equation}
\mbf{d} = \mbf{t} + \widetilde{\mbs{R}} \mbf{u} \mbs{R} - \mbf{v}, \quad \mbf{u} = \frac{1}{w} \sum_i w_i \mbf{u}_i, \quad \mbf{v} = \frac{1}{w} \sum_i w_i \mbf{v}_i, \quad w = \sum_i w_i.
\end{equation}
Isolating for $\mbf{t}$ and inserting this into~\eqref{eq:prob1} we have a new problem,
\begin{equation}\label{eq:prob2}
\min_{\mbs{R},\mbf{d}} \left( \sum_i w_i \left( (\mbf{u}_i - \mbf{u}) - \mbs{R} (\mbf{v}_i - \mbf{v} )\widetilde{\mbs{R}} \right)^2 + w \mbf{d}^2 \right),
\end{equation}
where the rotation and new translation variable, $\mbf{d}$, decouple.  We can minimize the second term with $\mbf{d} = \mbf{0}$ whereupon the optimal translation is
\begin{equation}\label{eq:transsoln}
\mbf{t} = \mbf{v} - \widetilde{\mbs{R}} \mbf{u} \mbs{R},
\end{equation}
in terms of the optimal rotation, for which we still must solve.  We are therefore now interested in solving
\begin{equation}\label{eq:prob3}
\min_{\mbs{R}}  \sum_i w_i \left( (\mbf{u}_i - \mbf{u}) - \mbs{R} (\mbf{v}_i - \mbf{v} )\widetilde{\mbs{R}} \right)^2,
\end{equation}
which is Wahba's classic rotation-only problem \citep{wahba65}.  

We notice that since we will enforce $\mbs{R}\widetilde{\mbs{R}} = \widetilde{\mbs{R}} \mbs{R}= 1$, it is possible to write the squared expression as
\begin{eqnarray}
\left( (\mbf{u}_i - \mbf{u}) - \mbs{R} (\mbf{v}_i - \mbf{v} )\widetilde{\mbs{R}} \right)^2 & = & \left( (\mbf{u}_i - \mbf{u}) - \mbs{R} (\mbf{v}_i - \mbf{v} )\widetilde{\mbs{R}} \right)\left( (\mbf{u}_i - \mbf{u}) - \mbs{R} (\mbf{v}_i - \mbf{v} )\widetilde{\mbs{R}} \right) \nonumber \\ & = & (\mbf{u}_i - \mbf{u})^2 + \mbs{R} (\mbf{v}_i - \mbf{v} ) \underbrace{\widetilde{\mbs{R}} \mbs{R}}_{1} (\mbf{v}_i - \mbf{v}) \widetilde{\mbs{R}} \nonumber \\ & & \hspace{1.0in} - (\mbf{u}_i - \mbf{u}) \mbs{R} (\mbf{v}_i - \mbf{v} ) \widetilde{\mbs{R}}  - \mbs{R} (\mbf{v}_i - \mbf{v} ) \widetilde{\mbs{R}} (\mbf{u}_i - \mbf{u})  \\ & = & (\mbf{u}_i - \mbf{u})^2 + (\mbf{v}_i - \mbf{v} )^2 - 2\left< \mbs{R} (\mbf{v}_i - \mbf{v}) \widetilde{\mbs{R}} (\mbf{u}_i - \mbf{u} ) \right> . \nonumber 
\end{eqnarray}
Therefore, the solution of
\begin{equation}\label{eq:prob4}
\max_{\mbs{R}}  \sum_i w_i \left< \mbs{R} (\mbf{v}_i - \mbf{v}) \widetilde{\mbs{R}} (\mbf{u}_i - \mbf{u} ) \right>,
\end{equation}
will have the same solution as~\eqref{eq:prob3}.  

\subsection{Davenport's Solution}

To enforce the rotor constraint, we introduce a Lagrange multiplier term and seek to solve the problem
\begin{equation}\label{eq:prob5}
\max_{\mbs{R}, \lambda} \biggl( \underbrace{\sum_i w_i \left< \mbs{R} (\mbf{v}_i - \mbf{v}) \widetilde{\mbs{R}} (\mbf{u}_i - \mbf{u} ) \right>}_{\rm benefit~function} + \lambda \underbrace{\left( 1- \left< \mbs{R} \widetilde{\mbs{R}} \right> \right)}_{\rm rotor~constraint} \biggr) ,
\end{equation}
where $\lambda$ is our Lagrange multiplier.

Taking the multivector derivative of~\eqref{eq:prob5} with respect to $\mbs{R}$ (considering it as a generic even multivector, not yet a rotor) and setting to zero we have the following condition (along with the rotor constraint) for extrema:
\begin{equation}\label{eq:extremaconds}
\sum_i w_i  (\mbf{v}_i - \mbf{v}) \widetilde{\mbs{R}} (\mbf{u}_i - \mbf{u} ) = \lambda \widetilde{\mbs{R}},
\end{equation}
where we made use of~\eqref{eq:multideriv} and Leibniz's product rule of differentiation.  We note that this equation is expressed strictly within $\mathbb{G}^{3^+}$, the subalgebra of even multivectors.  With $\mbs{R} = a + \mbs{I}\mbf{c}$, we can then define the following four even multivectors:
\beqn{}
x_0 - \mbs{I} \mbf{y}_0 & = & \sum_i w_i (\mbf{v}_i - \mbf{v}) (\mbf{u}_i - \mbf{u}), \\
x_1 - \mbs{I} \mbf{y}_1 & = & -\sum_i w_i (\mbf{v}_i - \mbf{v}) \mbf{e}_2 \mbf{e}_3 (\mbf{u}_i - \mbf{u}), \\
x_2 - \mbs{I} \mbf{y}_2 & = & -\sum_i w_i (\mbf{v}_i - \mbf{v}) \mbf{e}_3 \mbf{e}_1 (\mbf{u}_i - \mbf{u}), \\
x_3 - \mbs{I} \mbf{y}_3 & = & -\sum_i w_i (\mbf{v}_i - \mbf{v}) \mbf{e}_1 \mbf{e}_2 (\mbf{u}_i - \mbf{u}),
\eeqn
which can be constructed entirely from the data; there are actually only $10$ unique scalar variables summarizing all the data since we will end up with a symmetric $4 \times 4$ matrix.  With these definitions, we can rewrite the extrema condition in~\eqref{eq:extremaconds} as
\begin{equation} \label{eq:extremaconds2}
a \left( x_0 - \mbs{I} \mbf{y}_0 \right) + c_1 \left( x_1 - \mbs{I} \mbf{y}_1 \right) + c_2 \left( x_2 - \mbs{I} \mbf{y}_2 \right) + c_3 \left( x_3 - \mbs{I} \mbf{y}_3 \right) = \lambda \left( a - \mbs{I} \mbf{c} \right),
\end{equation}
where $\mbf{c} = (c_1, c_2, c_3)$.  Then, comparing the scalar and bivector components separately, we can write this as
\begin{equation}
\underbrace{\bbm x_0 & x_1 & x_2 & x_3 \\ \mbf{y}_0 & \mbf{y}_1 & \mbf{y}_2 & \mbf{y}_3 \ebm}_{\mbf{K}} \bbm a \\ \mbf{c} \ebm = \lambda \bbm a \\ \mbf{c} \ebm,
\end{equation}
a four-dimensional eigenproblem expressed in regular matrix algebra, which can be solved easily using numerical methods or one of the existing closed-form solutions \citep{shuster81,mortari97,yang13}.  In this form, the $\mbf{K}$ matrix can also be written as
\begin{equation}\label{eq:datamatrix}
\mbf{K} = \bbm \mbox{tr}(\mbf{Z}) & \mbf{x}^T \\ \mbf{x} & \mbf{Z} + \mbf{Z}^T - \mbox{tr}(\mbf{Z}) \ebm, \quad \mbf{Z} = \sum_i w_i (\mbf{v}_i - \mbf{v}) (\mbf{u}_i - \mbf{u})^T, \quad \mbf{x} = \bbm x_1 \\ x_2 \\ x_3 \ebm = \mbf{y}_0 = \bbm z_{23} - z_{32} \\ z_{31} - z_{13} \\ z_{12} - z_{21} \ebm, 
\end{equation}
which matches the familiar form of the `q-method' \citep{davenport65}; as is hopefully clear from the context, the multiplications in~\eqref{eq:datamatrix} use regular matrix algebra not the geometric product of \ac{GA}.

Returning to~\eqref{eq:extremaconds}, we can premultiply both sides by $\mbs{R}$ to see that
\begin{equation}
\mbs{R} \sum_i w_i  (\mbf{v}_i - \mbf{v}) \widetilde{\mbs{R}} (\mbf{u}_i - \mbf{u} ) = \lambda \underbrace{\mbs{R} \widetilde{\mbs{R}}}_{1},
\end{equation}
so that we can say
\begin{equation}
\sum_i w_i  \left< \mbs{R} (\mbf{v}_i - \mbf{v}) \widetilde{\mbs{R}} (\mbf{u}_i - \mbf{u} ) \right> = \lambda,
\end{equation}
at an extremum.  Comparing this to~\eqref{eq:prob5}, we see that the benefit function that we are maximizing is simply equal to $\lambda$ at an extremum.  This means that in solving the eigenproblem, we should choose the largest eigenvalue (along with its associated eigenvector) to maximize the benefit function.  To enforce the rotor constraint, we need only make the eigenvector unit length. Finally, once we have the optimal rotation, $\mbs{R} = a + \mbs{I} \mbf{c}$, we can return to~\eqref{eq:transsoln} to compute the optimal translation, $\mbf{t}$.

\subsection{Including a Rotation Prior and/or Measurements}

In addition to the pointcloud measurements, we may have one or more direct measurements of the rotor, $\mbs{S}_j$, or equivalently a prior rotation.  To keep our problem quadratic in the unknown rotor, we can modify~\eqref{eq:prob5} to be
\begin{equation}\label{eq:prob6}
\max_{\mbs{R}, \lambda} \biggl( \underbrace{\sum_i w_i \left< \mbs{R} (\mbf{v}_i - \mbf{v}) \widetilde{\mbs{R}} (\mbf{u}_i - \mbf{u} ) \right>}_{\rm point~measurements} + \underbrace{\sum_j w_j \left<  \mbs{R}^2 \widetilde{\mbs{S}}_j^2 \right>}_{\rm rotation~measurements} + \lambda \underbrace{\left( 1- \left< \mbs{R} \widetilde{\mbs{R}} \right> \right)}_{\rm rotor~constraint} \biggr).
\end{equation}
The derivative of the new benefit term is
\begin{equation}
\frac{\partial}{\partial \mbs{R}} \sum_j w_j \left<  \mbs{R}^2 \widetilde{\mbs{S}}_j^2 \right> = \mbs{R} \underbrace{ \sum_j w_j  \widetilde{\mbs{S}}_j^2 }_{g - \mbs{I} \mbf{h}} + \underbrace{ \sum_j w_j  \widetilde{\mbs{S}}_j^2 }_{g - \mbs{I} \mbf{h}} \mbs{R} = 2 (a g + \mbf{h}\cdot \mbf{c}) - 2 \mbs{I} ( a \mbf{h} - g \mbf{c} ),
\end{equation}
where we convert the weighted sum of (reversed) rotor measurements into $g - \mbs{I} \mbf{h}$, another even multivector.  Including this into the extrema conditions in~\eqref{eq:extremaconds2} we have
\begin{equation}
a \left( x_0 - \mbs{I} \mbf{y}_0 \right) + c_1 \left( x_1 - \mbs{I} \mbf{y}_1 \right) + c_2 \left( x_2 - \mbs{I} \mbf{y}_2 \right) + c_3 \left( x_3 - \mbs{I} \mbf{y}_3 \right) + (a g + \mbf{h}\cdot \mbf{c}) - \mbs{I} ( a \mbf{h} - g\mbf{1} \mbf{c} )= \lambda \left( a - \mbs{I} \mbf{c} \right).
\end{equation}
Comparing the scalar and bivector terms, the eigenproblem becomes
\begin{equation}
\underbrace{\bbm \mbox{tr}(\mbf{Z}) + g & (\mbf{x} + \mbf{h})^T \\ \mbf{x} + \mbf{h}  & \mbf{Z} + \mbf{Z}^T - \mbox{tr}(\mbf{Z}) - g\mbf{1} \ebm}_{\mbf{K}} \bbm a \\ \mbf{c} \ebm = \lambda \bbm a \\ \mbf{c} \ebm,
\end{equation}
where we have a modified $\mbf{K}$ matrix.  From here we can proceed as before.

When there are no pointcloud measurements, only the direct measurements of the rotor, the eigenproblem becomes
\begin{equation}
\bbm  g & \mbf{h}^T \\ \mbf{h}  & - g\mbf{1} \ebm \bbm a \\ \mbf{c} \ebm = \lambda \bbm a \\ \mbf{c} \ebm.
\end{equation}
Writing the second condition out in full we see that
\begin{equation}
a \mbf{h} - g \mbf{c} = \lambda \mbf{c},
\end{equation}
so it must be that $\mbf{c}$ is parallel to $\mbf{h}$.  Letting $\mbf{c} = c \mbf{n}$ and $\mbf{h} = h \mbf{n}$, with $\mbf{n}$ a unit vector in the direction of $\mbf{h}$ the eigenproblem collapses to
\begin{equation}
\bbm  g & h \\ h  & - g\ \ebm \bbm a \\ c \ebm = \lambda \bbm a \\ c \ebm.
\end{equation}
The characteristic equation is
\begin{equation}
\lambda^2 - (g^2 + h^2) = 0.
\end{equation}
The unit eigenvector associated with the largest eigenvalue, $\lambda = \sqrt{ g^2 + h^2 }$, is
\begin{equation}
\bbm a \\ c \ebm = \frac{1}{\sqrt{2\left(g^2 + 2 g \sqrt{g^2 + h^2} + h^2 \right)}} \bbm \sqrt{ g^2 + h^2 } + g \\ h \ebm. 
\end{equation}
The optimal rotor is $\mbs{R} = a + c \mbs{I} \mbf{n}$.

%\newpage
\bibliographystyle{asrl}
\bibliography{refs,refs2,book,pubs,refs_ga}

\end{document}